\documentclass[conference]{IEEEtran} 
\IEEEoverridecommandlockouts
\usepackage{cite}
\usepackage{amsmath,amssymb,amsfonts}
\usepackage{algorithm}
\usepackage{algorithmic}
\usepackage{graphicx}
\usepackage{textcomp}
\usepackage{xcolor}
\usepackage{booktabs}
\def\BibTeX{{\rm B\kern-.05em{\sc i\kern-.025em b}\kern-.08em
    T\kern-.1667em\lower.7ex\hbox{E}\kern-.125emX}}
\begin{document}

\title{
Runtime Safety Filtering for Learned Small UAS Separation Policies under GNSS Degradation

}

\author{\IEEEauthorblockN{Alex Zongo \qquad Peng Wei}
\IEEEauthorblockA{\textit{Department of Mechanical and Aerospace Engineering} \\
\textit{George Washington University}\\
Washington, D.C., USA \\
\tt\small\{a.zongo, pwei\}@gwu.edu.}
}


\maketitle

\begin{abstract}

Learning-based separation assurance, notably reinforcement learning, for small Unmanned Aircraft Systems (sUAS) has demonstrated strong performance in simulation, achieving near-zero collision rates while maintaining traffic throughput in high-density scenarios. However, these learned control policies assume accurate position and velocity information derived from Global Navigation Satellite Systems (GNSS); an assumption that fails in urban environments where multipath propagation, signal blockage, and intentional interference routinely degrade navigation integrity. Under such conditions, when deploying a learned aircraft separation assurance policy, a fundamental architectural question arises: should runtime safety mechanisms filter the policy's actions to satisfy some constraints, or filter its observation input to present a conservative view of the traffic state? This work evaluates both approaches for multi-agent sUAS separation under adversarial GNSS degradation. Both architectures share a common first step: estimating a worst-case traffic state consistent with some bounded observation uncertainty. The approaches differ thereafter. Action filtering constrains policy outputs to satisfy hand-designed safety conditions, implemented via discrete-time control barrier functions, and evaluated at the worst-case state. Observation filtering, on the other hand, presents the worst-case state to the policy as a corrected input, allowing the policy to determine its own response. 
Experimental evaluation reveals that action filtering provides negligible safety improvement, while observation filtering reduces near mid-air collisions by 90\% and remains robust to the control barrier function's tradeoff between separation distance and closing rate. These results suggest that, for policies with learned safety behaviors, preserving the policy's decision authority outperforms overriding its actions with hand-designed control filtering. 

\end{abstract}


\section{Introduction}
\label{sec:intro}

The integration of small unmanned aircraft systems (sUAS) into low-altitude urban airspace presents a fundamental challenge: tactical separation assurance must scale beyond human-supervised operations while maintaining safety standards that aviation demands. Under the UAS Traffic Management (UTM) paradigm \cite{faa_utm, utm_concept}, high-density traffic flows require decentralized deconfliction, where each aircraft makes autonomous speed or heading adjustments to maintain safe separation from nearby traffic.
Multi-agent reinforcement learning (MARL) has emerged as a promising approach to this challenge, with recent work demonstrating separation policies that achieve near-zero collision rates while maintaining efficient traffic flow in simulation \cite{brittain1, brittain2, shulu, arsyi}. These policies learn sophisticated deconfliction behaviors through millions of training interactions, developing responses/strategies to multi-aircraft encounter scenarios that would be difficult to hand-design. 

A critical assumption underlies this success: policies receive accurate aircraft state information derived from Global Navigation Satellite Systems (GNSS). In urban environments, this assumption often fails. Multipath propagation from building surfaces introduces position errors exceeding tens of meters \cite{misra, peretic}. Signal blockage in urban canyons creates coverage gaps. Intentional interference, whether it is jamming or spoofing, poses an increasing threat to unmanned operations \cite{psiaki, gutierrez, sathaye, kerns}. When a separation control policy receives degraded observations, it may misinterpret encounter geometries: given two sUAS where one of them, considered as an intruder, is on a collision course with the other, it may appear safely separated, or closing velocities may seem divergent when they are not. This results in a degraded safety performance of separation assurance, and robust behavior is needed.  

Recent work has addressed this challenge through adversarial training, developing policies that maintain safety under bounded observation perturbations \cite{zongo_robust_marl}. However, while training-time robustness provides probabilistic guaranties, operational deployment may encounter degradation levels or patterns beyond training assumptions. A complementary approach is runtime safety filtering: mechanisms that intervene during execution to prevent unsafe outcomes regardless of training history. The question that this paper addresses is how to design such mechanisms to work effectively with learned policies. 

Runtime safety filtering introduces an architectural choice. Given degraded observations, one can compute worst-case states consistent with the uncertainty. Then we either (\emph{i}) constrain the policy's actions to satisfy some safety conditions evaluated at that worst-case state, or (\emph{ii}) correct the policy's observations by presenting the worst-case state as input. Both approaches use identical worst-case computations; they differ only in what follows after that. This paper evaluates both architectures for sUAS separation under adversarial GNSS degradation. To that end, the contributions are as follows.

\begin{enumerate}
    \item An empirical comparison demonstrates that observation filtering reduces near mid-air collisions effectively, while action filtering provides negligible improvement despite using identical worst-case state estimation. 
    \item A sensitivity analysis reveals that observation filtering is robust to the closing rate coefficient in the designed control barrier function \eqref{eq:barrier}, whereas action filtering performance depends critically on how the safety constraint is formulated.  
    \item An explanation of why preserving the policy's decision authority outperforms overriding its actions with hand-designed constraints, informing the design of runtime safety mechanisms for learned controllers.
\end{enumerate}


\section{Related Work}
\label{sec:related_works}

\subsection{Robust Reinforcement Learning}
\label{subsec:robust_rl}

Training policies to withstand observation perturbations and degradation has received substantial attention in the reinforcement learning (RL) community. Zhang et al. \cite{zhang2020robust} formulate state-adversarial Markov Decision Processes (MDP) where an adversary perturbs observations within bounded sets, and training policies via alternating optimization between policy improvement and adversarial attack. Pinto et al. \cite{pinto2017robust} apply similar ideas to robotic manipulation, demonstrating improved transfer from simulation to physical systems. For multi-agent separation assurance specifically, Zongo et al. \cite{zongo_robust_marl} develop an $R$-contamination observation model with closed-form adversarial perturbations derived from value-function gradients, avoiding the instability of learned adversaries while providing theoretical bounds on performance degradation. Although these methods address training-time robustness, our work considers runtime mechanisms that complement or substitute for such training. 

\subsection{Control Barrier Functions under Uncertainty}
\label{subsec:cbf_under_uncertainty}

Control Barrier Functions (CBFs) provide formal safety guarantees by constraining control inputs to maintain forward invariance of safe sets \cite{ames2017cbf}. Given a set of safe states defined by $\{ x: h(x) \ge 0 \}$ for some continuously differentiable function $h$, the barrier condition ensures that trajectories starting in the safe set remain there. Under state uncertainty, robust CBF formulations tighten constraints to account for worst-case states within uncertainty sets \cite{jankovic2018robust}. Dean et al. \cite{dean2020guaranteeing} extend this to learned perception modules, deriving measurement-robust CBFs that guarantee safety despite bounded perception errors. Moreover, Cosner et al. \cite{cosner2023robust} address model uncertainty with Lipschitz bounds. While these approaches filter actions to satisfy safety constraints, we additionally consider filtering observations as an alternative architecture.

\subsection{Runtime Safety for Learned Controllers}
\label{subsec:runtime_safety}

Integrating learned policies with runtime safety has motivated several architectures. 
Fisac et al. \cite{fisac2019bridging} use Hamilton-Jacobi reachability analysis to provide safety guarantees for learning systems, computing backward reachable sets that define when intervention, for example by taking a specific action, is necessary. Cheng et al. \cite{cheng2019end} learn CBF parameters jointly with policy training, enabling end-to-end optimization of both performance and safety. Thananjeyan et al. \cite{thananjeyan2021filtering} learn filtering policies that are activated when the primary policy approaches unsafe states. Our work differs in evaluating two uses of the same worst-case state computation, i.e., filtering actions versus filtering observations for a fixed pre-trained policy, isolating the architectural choice from other design decisions.

\section{Problem Formulation}
\label{sec:problem_formulation}

\subsection{Aircraft Separation Assurance}

We consider the multi-agent separation assurance problem formulated in \cite{zongo_robust_marl}. A set of $N$ small UASs operate in a shared en route urban airspace, each following a predefined route where tactical deconfliction is achieved by speed adjustments. Each aircraft $i$ is described by a state vector $s_i=(x_i, y_i, \psi_i, v_i, d_{g,i}, a_{i}^{-})$ comprising planar positions ($x_i, y_i$), heading ($\psi_i$), speed ($v_i$), distance to goal or destination ($d_{g,i}$) and previous action ($a_i^-$). The heading evolves according to the route or the flight plan, while the speed is controlled by the separation policy. 
Moreover, each agent or sUAS not only observes its own state but also receives broadcast states of nearby traffic via Remote ID \cite{faa_remoteID}, forming observation $o_i$ that includes positions, headings, and speeds of intruders within sensing range.
Furthermore, the separation standard is $D_{\text{safe}} = 100$ m horizontal \cite{zongo_robust_marl}. A near mid-air collision (NMAC) occurs when two aircraft pass within distance $D_{\text{safe}}$. The goal of aircraft separation assurance is to eliminate or minimize the number of NMACs in a given airspace. 

The action space is discrete: $a \in \{ -1, \ 0, \ 1 \}$, i.e., corresponding to decelerate ($-1$), hold ($0$), or accelerate ($1$). Each action modifies the ground speed by increment $\Delta v \approx 2.57$ m/s \cite{brittain1, faa_speed_adjust, zongo_robust_marl} per decision time step. This discrete structure reflects certification-oriented avionics integration where high-level autonomy issues bounded setpoints to flight control systems. 

\begin{figure}[t]
    \centering
    \includegraphics[width=\linewidth]{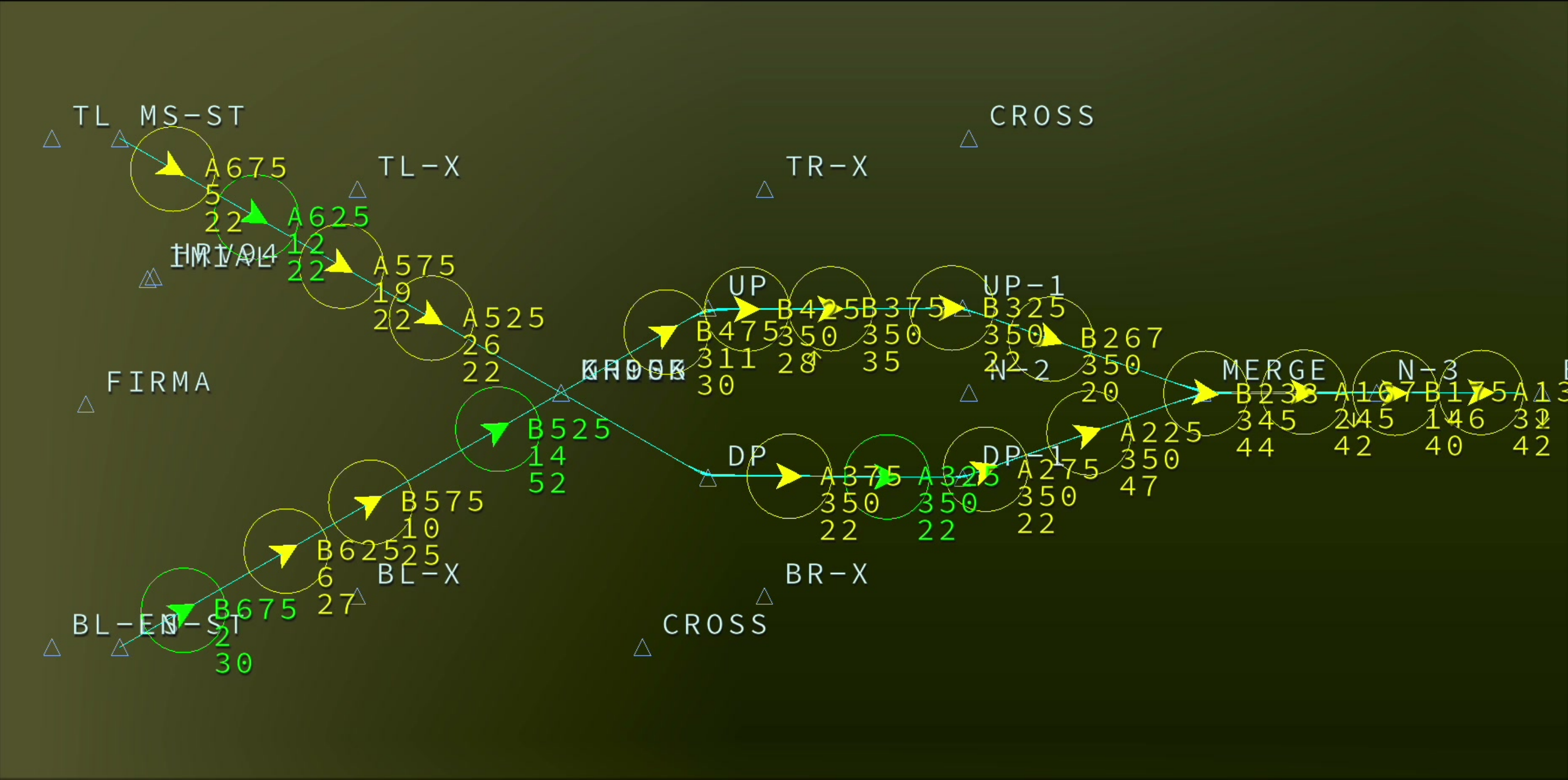}
    \caption{Snapshot of the high-density en route urban airspace sUAS traffic simulation environment via the BlueSky Air Traffic Simulator \cite{hoekstra2016bluesky}. Each route spans about $10$ km, illustrating a compact, realistic scenario. Figure from \cite{zongo_robust_marl}.}
    \label{fig:simulation_env}
\end{figure}

Fig. \ref{fig:simulation_env} illustrates the geometry and simulation of a representative en route urban airspace. 

\subsection{Trained Policy}
\label{sec:pretrained_policy}

We use a separation control policy trained via Proximal Policy Optimization \cite{schulman2017ppo} under nominal (undegraded) observations, following the training framework of \cite{zongo_robust_marl}. The policy $\pi(a \mid o)$ maps state observations to a distribution over possible actions (e.g., slow down, hold speed, or accelerate). Under nominal conditions (i.e., when GNSS observations accurately reflect the traffic state), the policy achieves near-zero NMAC rates while maintaining air traffic throughput. 

Therefore, the trained policy exhibits learned safety behaviors: anticipatory responses to aircraft encounters, smooth speed adjustment sequences, and implicit coordination with nearby traffic. These behaviors emerge from reward shaping that penalizes separation violations and rewards efficient route completion, rather than explicit constraint satisfaction. The policy parameters encode implicit safety knowledge developed through the millions of training interactions within the simulated environment \cite{zongo_robust_marl}.

\subsection{GNSS Degradation Model}
\label{sec:gnss_degradation}

At deployment, GNSS degradation or spoofing can significantly hinder position and velocity estimates. We model this degradation using a deterministic formulation that captures continuous signal deterioration. 

\begin{equation}
    o = (1-R) s + R\xi, \quad \| \xi - s\|_\infty \leq \kappa,
    \label{eq:r_contam_model}
\end{equation}
where $s$ is the true state, $o$ is the observed (degraded) state, $R\in [0,1)$ represents the degradation intensity. $\xi$ is the adversarial perturbation and $\kappa$ bounds the perturbation magnitude componentwise. 

The parameter $R$ quantifies GNSS degradation severity. At $R=0$, observations equal true states. As $R$ increases, observations become increasingly degraded, approaching complete adversarial perturbation as $R \to 1$. This parameter captures the continuum from nominal GNSS performance to severe multipath or partial spoofing. 

Moreover, adversarial perturbation $\xi$ represents the worst-case degradation within the bounded set. Rather than modeling benign noise, we assume an adversary that selects $\xi$ to maximally obscure danger: inflating observed separation distances and making closing velocities appear divergent. This conservative assumption stress-tests safety mechanisms against intentional manipulation and subsumes less adversarial perturbation modes. 

\begin{figure}[t]
    \centering
    \includegraphics[width=0.65\linewidth]{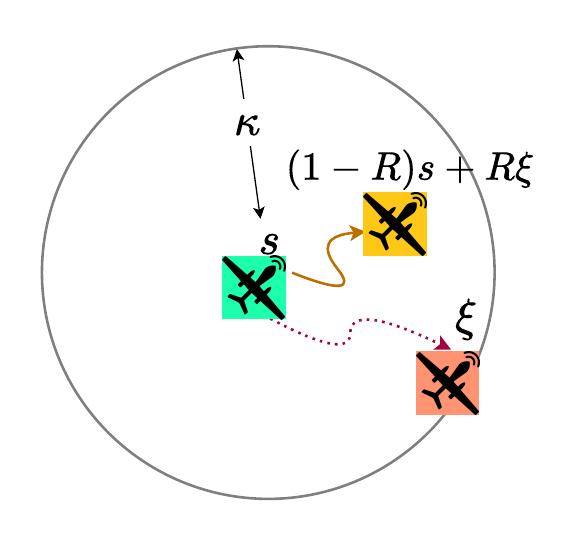}
    \caption{Deterministic model for GNSS degradation. The true sUAS state $s$ (green) is degraded to observed state $o=(1-R)s + R\xi$ (yellow), that is convex combination with adversarial perturbation $\xi$ (red). The perturbation is bounded by $\| \xi - s \|_\infty\le \kappa $, yielding an uncertainty set of radius $R\kappa$ around the observation.}
    \label{fig:gnss_model}
\end{figure}

Fig. \ref{fig:gnss_model} illustrates the GNSS degradation model. The observed state lies on the segment between true state and adversarial perturbation, whose position is determined by $R$. 

This formulation differs from the probabilistic model used during adversarial training in \cite{zongo_robust_marl}, where observations are equal to the true state with probability $1-R$ and an adversarial state with probability $R$. Although this probabilistic model suits expected value optimization during training, this deterministic GNSS degradation model provides the bounded uncertainty sets required for worst-case runtime analysis.  

Furthermore, by inverting \eqref{eq:r_contam_model}, the true state given observation $o$ lies within the $\ell_\infty$ ball or uncertainty set:
\begin{equation}
    \mathcal{U}(o, R) = \left\{ s: \| s - o\|_\infty \leq R\kappa \right\}, \quad \delta \triangleq R\kappa,
    \label{eq:true_state_uncertainty_set}
\end{equation}
enabling worst-case computation for runtime safety filtering. 

In practice, the position and speed estimates of each sUAS are subject to GNSS degradation with bounds $\kappa_x = \kappa_y = 60$ m and $\kappa_v = 2$ m/s. Heading is derived from magnetometer and inertial sensors and is assumed to be correct.

\subsection{Runtime Safety Objective}

Given a policy $\pi$ trained under nominal conditions and given degraded observation $o$ at deployment, we seek runtime mechanisms that reduce near mid-air collisions (NMAC) rates across degradation levels $R$ while preserving effectiveness when degradation is absent. The mechanism should require no retraining of the underlying policy and add minimal computational overhead to real-time operation.

\section{Control Barrier Function Preliminaries}
\label{sec:control_barrier_function_preliminaries}

\subsection{Safety as Set Invariance}

Consider a dynamical system with state $s \in \mathbb{R}^n$ and control input $a \in \mathcal{A}$. Safety can be formalized as maintaining the state within a designated safe set $\mathcal{C} \subset \mathbb{R}^n$. If the system starts in $\mathcal{C}$ and remains there for all future time, we say $\mathcal{C}$ is \emph{forward invariant} under the control policy $\pi(s)=a$ \cite{ames2017cbf}. 

For separation assurance, the safe set can be defined by maintaining minimum separation between an sUAS $i$ and all other nearby aircraft $j$:
\begin{equation}
    \mathcal{C} = \{ s: \min_j\| p_i-p_j \| \geq D_{\text{safe}} \},
\end{equation}
where $p_k$ denotes the position of agent $k$. The runtime safety problem is to ensure forward invariance of $\mathcal{C}$ despite GNSS-induced observation degradation. 

\subsection{Barrier Functions}

A barrier function $h: \mathbb{R}^n \to \mathbb{R}$ encodes the safe set as its \emph{zero-superlevel} set: $\mathcal{C}=\{ s: h(s) \geq 0 \}$. The function $h(s)$ measures the safety margin, where positive values indicate safety, negative values indicate safety violation, and zero marks the boundary. 

For separation assurance, we design the barrier function $h$ to incorporate both separation distance and closing rate:
\begin{equation}
    h(s) = (\|r\|^2 - D_{\text{safe}}^2) + \alpha \cdot r^\top v_{\text{rel}},
    \label{eq:barrier}
\end{equation}
where $r = p_i - p_j$ is the relative position with $p_i=[x_i, y_i]^\top$ and $p_j=[x_j, y_j]^\top$. $v_{\text{rel}} = v_i d_i - v_j d_j$ is the relative velocity with $d_k=[\cos\psi_k, \sin\psi_k]^\top$ the heading unit vector, and $\alpha > 0$ (in seconds) weights the closing rate term.

The first term, $\|r\|^2 - D_{\text{safe}}^2$ in \eqref{eq:barrier}, is positive when separation exceeds $D_{\text{safe}}$. The second term, $\alpha \cdot r^\top v_{\text{rel}}$, is positive when, for example, two sUASs are diverging and negative when they close on each other. The parameter $\alpha$ balances these contributions: a larger $\alpha$ makes the barrier more sensitive to the closing rate.

\subsection{Discrete-Time CBF Condition}
\label{sec:discrete_cbf_prelim}

Our simulation environment evolves in discrete time with a timestep of $1$ s. Let $s_k = f(s_k, a_k)$ denote the state transition under action $a_k$. Following the discrete-time CBF formulation of \cite{agrawal2017discrete}, forward invariance is ensured by requiring:
\begin{equation}
    h(s_{k+1}) \geq (1-\gamma) h(s_k)
    \label{eq:dtcbf}
\end{equation}
where $\gamma\in(0,1)$ bounds the decay rate per-step of the barrier function. 

This condition guarantees that if $h(s_k) \geq 0$, then $h(s_{k+1}) \geq (1-\gamma) h(s_k) \geq 0$, hence maintaining safety. The parameter $\gamma$ controls conservatism: smaller $\gamma$ allows slower barrier decay, allowing the system to approach the safety limit more gradually; larger $\gamma$ enforces more aggressive safety margins.

\section{Runtime Safety Filters}
\label{sec:runtime_safety_approaches}

We evaluate two runtime filtering approaches that share a common first step, that is, worst-case true state estimation, but differ in how this information is subsequently used. 

\subsection{Worst-Case State Estimation}

Given a degraded observation $o$ and an assumed GNSS degradation level $R_{\text{max}}$, the worst-case true state minimizes the barrier over the uncertainty set \eqref{eq:true_state_uncertainty_set} which yields:
\begin{equation}
    s^\star = \arg\min_{s\in \mathcal{U}(o,R_{\text{max}})} h(s).
    \label{eq:worst_case}
\end{equation}
The worst-case state $s^\star$ represents the most dangerous local traffic state consistent with the degraded observation: minimum separation with maximum closing rate.

The assumed bound $R_{\text{max}}$ is a design parameter that reflects the expected worst GNSS quality in operational environment. It does not need to equal the true degradation level $R$, which is unknown at runtime. 

Furthermore, the minimization \eqref{eq:worst_case} admits a closed-form solution. In fact, for positions, the quadratic and linear terms of the barrier function are jointly minimized by displacing the positions of a pair of sUASs ($i,j$) towards each other along the line of sight. Let $\tilde{p}_i$, and $\tilde{p}_j$ denote the observed (possibly degraded) position of sUASs $i$ and $j$. And $\delta_p = R_{\text{max}} \kappa_p$, which represents the designed uncertainty radius of the position estimation. Thus, 
\begin{align}
    p_i^\star &= \tilde{p}_i - \delta_p \cdot \text{sign}(\tilde{r}), \label{eq:pos_own} \\
    p_j^\star &= \tilde{p}_j + \delta_p \cdot \text{sign}(\tilde{r}) ,
    \label{eq:pos_int}
\end{align}
where $\tilde{r} = \tilde{p}_i - \tilde{p}_j$ and $\text{sign}(\cdot)$ operate componentwise. 

Then we estimate the worst-case speed for the pairs of sUASs ($i, j$), minimizing the additional linear term $-r^\top v_{\text{rel}}$ that depends on $v_i$ and $v_j$. This translates into maximizing the closing rate $r^\top v_{\text{rel}}$ via:
\begin{align}
    v_i^\star &= \tilde{v}_i - \delta_v \cdot \text{sign}(r^{*\top} d_i), \label{eq:vel_own} \\
    v_j^* &= \tilde{v}_j + \delta_v \cdot \text{sign}(r^{*\top} d_j), \label{eq:vel_int}
\end{align}
where $r^\star = p_i^* - p_j^*$ and $\delta_v = R_{\max} \kappa_v$. 
Headings remain true at observed values.

\subsection{Action Filtering via Discrete-Time CBF}

Let us assume that the policy output is $a_{\text{nom}} = \pi(o)$. Action filtering uses the worst-case state defined in \eqref{eq:worst_case} to identify a safe action that satisfies \eqref{eq:dtcbf}. 

Given our discrete action space $\mathcal{A} = \{-1, 0, +1\}$, we enumerate all actions and identify those satisfying the constraint \eqref{eq:dtcbf} so that:
\begin{equation}
    \mathcal{A}_{\text{safe}}(s_k) = \{a \in \mathcal{A} : h(f(s_k, a)) \geq (1-\gamma) h(s_k)\}.
    \label{eq:safe_actions}
\end{equation}
If $\mathcal{A}_{\text{safe}} \neq \emptyset$, we select the feasible action closest to the policy's output. However, if $\mathcal{A}_{\text{safe}} = \emptyset$, i.e., no action satisfies the constraint, we select the action minimizing constraint violation. This enumeration is exact for small discrete action spaces and computationally trivial.

More precisely, for each candidate action $a \in \{ -1,0,+1 \}$ we predict the next-step barrier value $h(s_{k+1})=h_{k+1}$. An sUAS $i$'s speed change under action $a$ is $\Delta v \cdot a$ in the heading direction, yielding next-step relative position and velocity:
\begin{align}
    r_{k+1} &= r^* + \Delta t \cdot (v_{\text{rel}}^* + \Delta v \cdot a \cdot d_{i}), \\
    v_{\text{rel}, k+1} &= v_{\text{rel}}^* + \Delta v \cdot a \cdot d_i,
\end{align}
where $\Delta t = 1$ s is the time step. The predicted barrier value is defined as follows:
\begin{equation*}
    h_{k+1}(a) = \|r_{k+1}\|^2 - D_{\text{safe}}^2 + \alpha \cdot r_{k+1}^\top v_{\text{rel}, k+1}.
\end{equation*}

Algorithm \ref{alg:cbf_filter} summarizes the action filtering procedure.
 
\begin{algorithm}[t]
\caption{CBF Action Filtering}
\label{alg:cbf_filter}
\begin{algorithmic}[1]
\REQUIRE Observation $o$, policy $\pi_\theta$, parameters $R_{\max}$, $\gamma$
\STATE $a_{\text{nom}} \gets \pi_\theta(o)$ \COMMENT{Policy proposed action}
\STATE $s^* \gets \textsc{WorstCaseState}(o, R_{\max})$ \COMMENT{Eqs. (\ref{eq:pos_own})--(\ref{eq:vel_int})}
\STATE $h_k \gets h(s^*)$ \COMMENT{Current barrier value}
\STATE $\mathcal{A}_{\text{safe}} \gets \emptyset$
\FOR{$a \in \{-1, 0, +1\}$}
    \STATE $h_{k+1} \gets h(f(s^*, a))$ \COMMENT{Predicted barrier}
    \IF{$h_{k+1} \geq (1-\gamma) h_k$}
        \STATE $\mathcal{A}_{\text{safe}} \gets \mathcal{A}_{\text{safe}} \cup \{a\}$
    \ENDIF
\ENDFOR
\IF{$\mathcal{A}_{\text{safe}} \neq \emptyset$}
    \RETURN $\arg\min_{a \in \mathcal{A}_{\text{safe}}} |a - a_{\text{nom}}|$ \COMMENT{Closest safe action}
\ELSE
    \RETURN $\arg\max_{a \in \mathcal{A}} h(f(s^*, a))$ \COMMENT{Least-violating action}
\ENDIF
\end{algorithmic}
\end{algorithm}

\subsection{GNSS State Observation Filtering}

Observation filtering uses the worst-case state differently: rather than constraining the actions, it passes a corrected observation to the policy or controller. In fact, given observation $o$ and worst-case state $s^\star$, we construct the corrected observation $o^\star$ by replacing the position and speed components with their worst-case estimated true values from \eqref{eq:pos_own}---\eqref{eq:vel_int}:
\begin{equation}
    a = \pi(o^\star).
    \label{eq:obs_filtering}
\end{equation}
The policy's output is used directly without modification.

This approach relies on the policy to respond appropriately to any aircraft encounters. If the policy has learned that close proximity of two sUASs with a high closing rate warrants slowing down, it applies this response to the worst-case observation just as it would to ground truth.

\section{Experimental Setup}
\label{sec:experimental_setup}

We evaluate both methods using the BlueSky air traffic simulation environment \cite{hoekstra2016bluesky}, configured for urban sUAS operations on en route structured airspace following \cite{zongo_robust_marl}. The scenario comprises a structured airspace with crossing and merging routes representative of package-delivery traffic, generating approximately 65 aircraft per episode across various encounter scenarios. Each sUAS is modeled after the Amazon MK30 with cruise speed $20$ m/s and speed range $7.5-36$ m/s \cite{zongo_robust_marl}.

We study the effects of both strategies across GNSS degradation levels or intensities $R \in \{0, 0.05, 0.1, \ldots, 1.0\}$ with adversarial perturbation as described in Section \ref{sec:gnss_degradation}. For both approaches, we sweep through assumed uncertainty bounds $R_{\max} \in \{0.1, 0.4, 0.5, 0.7, 1.0\}$. We also perform sensitivity analyses by varying the velocity weighting coefficient $\alpha \in \{1, 5, 10, 20\}$ s from \eqref{eq:barrier}. The CBF decay parameter is $\gamma = 0.2$. Each configuration is evaluated across 100 episodes.

We report near mid-air collision (NMAC) rates per episode, where an NMAC occurs when the separation between two aircraft falls below $D_{\text{safe}}=100$ m. Moreover, we report the average of the minimum separation distance achieved across all pairwise encounters per episode. Furthermore, for action filtering, we measure the fraction of decisions where the filter modifies the policy's output, i.e., the action override rate. In addition, we report the infeasibility rate defined as the fraction of decisions where no action satisfies the CBF constraint. 

\textbf{Computational Overhead}. Both filters share the worst-case state estimation step whose closed-form solution is $O(M)$ per aircraft, where $M$ is the number of intruders within sensing range. Action filtering adds an enumeration over the three discrete actions, each requiring a single barrier evaluation, while observation filtering adds one policy forward pass on the corrected observation. Both approaches add negligle cost relative to the $1$ s decision step, given the decentralized settings where each agent onboard hardware handles the computation. Therefore, neither approach requires retraining or specialized hardware beyond what the baseline policy already uses. 

\section{Results}
\label{sec:results}

\subsection{Action Filtering}

\begin{figure}[t]
    \centering
    \includegraphics[width=\linewidth]{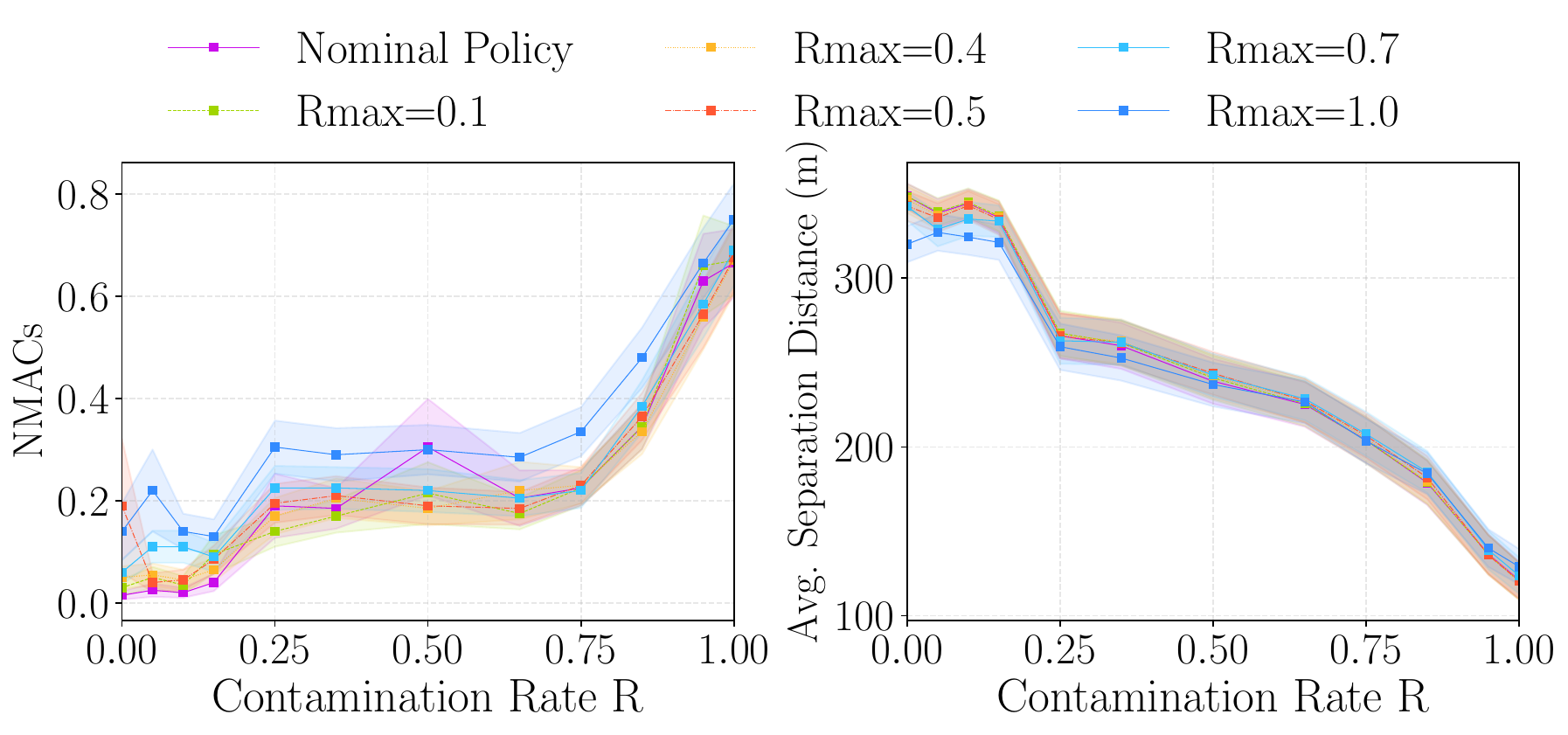}
    \caption{Safety performance of the action filter under increasing observation degradation. \emph{Left}: Near mid-air collision (NMAC) count of small UAS. \emph{Right}: Minimum separation distance between the aircraft agents achieved per episode. The performance remains relatively the same independently of the designed filter's $R_{\max}$. Shaded regions indicate $\pm$ standard error.}
    \label{fig:cbf_action_filter}
\end{figure}

Fig. \ref{fig:cbf_action_filter} presents the safety performance of the policy when coupled with action filtering across assumed uncertainty bounds $R_{\max}$. The results indicate that NMAC rates and minimum separation distances are statistically indistinguishable from nominal operation regardless of $R_{\max}$ setting. Therefore, we could argue that the action filter seems to provide negligible safety improvement. 

\begin{figure}[t]
    \centering
    \includegraphics[width=\linewidth]{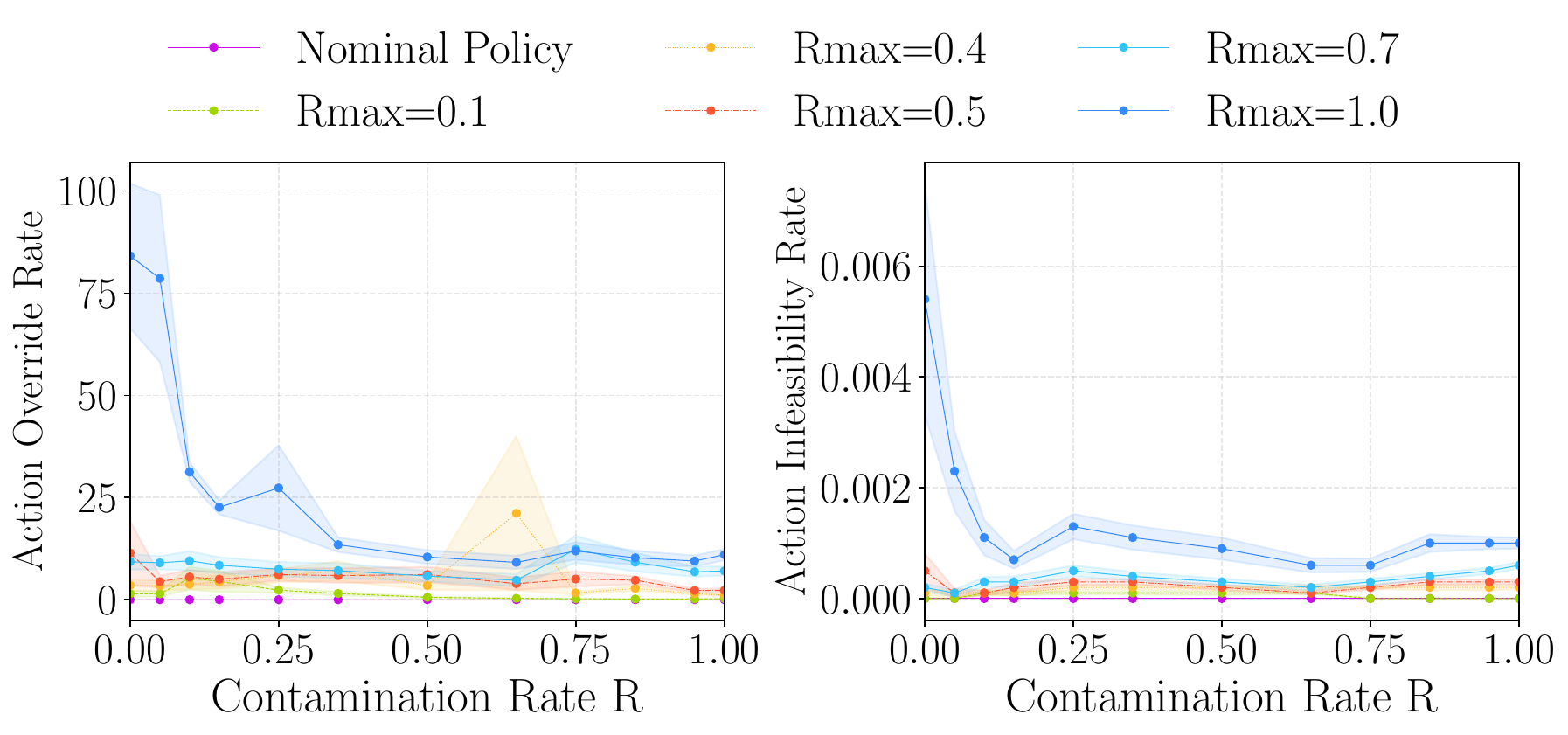}
    \caption{
    Action filter override and infeasibility rates. \emph{Left}: override rate remains relatively low for $R_{\max} \leq 0.7$, indicating the policy's actions typically satisfy the CBF constraint. At $R_{\max} = 1.0$, action overrides approach 100\% at low contamination. \emph{Right}: Infeasibility rate is negligible, confirming feasible actions always exist. Shaded regions indicate $\pm$ standard error. 
    }
    \label{fig:cbf_intervention_and_infeasibility_rate}
\end{figure}

Interestingly, Fig. \ref{fig:cbf_intervention_and_infeasibility_rate} appears to explain this behavior. The left panel shows the fraction of decisions where the filter modifies the policy's proposed action. For $R_{\max} \leq 0.7$, the action override rate is relatively low, though not negligible. For $R_{\max}=1.0$, it reaches $100\%$ at low degradation levels $R$. In other words, the maximally conservative worst-case state estimate sees danger everywhere, triggering constant overrides of policy actions. However, this aggressive intervention does not improve safety.  
The right panel shows infeasibility rates, i.e., the fraction of decisions where no action satisfies the CBF constraint \eqref{eq:dtcbf}. With infeasibility rates negligible across all configurations, the discrete action space always admits a constraint-satisfying action. Thus, the problem is not the absence of safe actions, but that the filter either does not activate, or it disrupts learned behaviors without improving safety outcomes.

\subsection{Observation Filtering}
\label{sec:obs_filtering}

\begin{figure}[t]
    \centering
    \includegraphics[width=\linewidth]{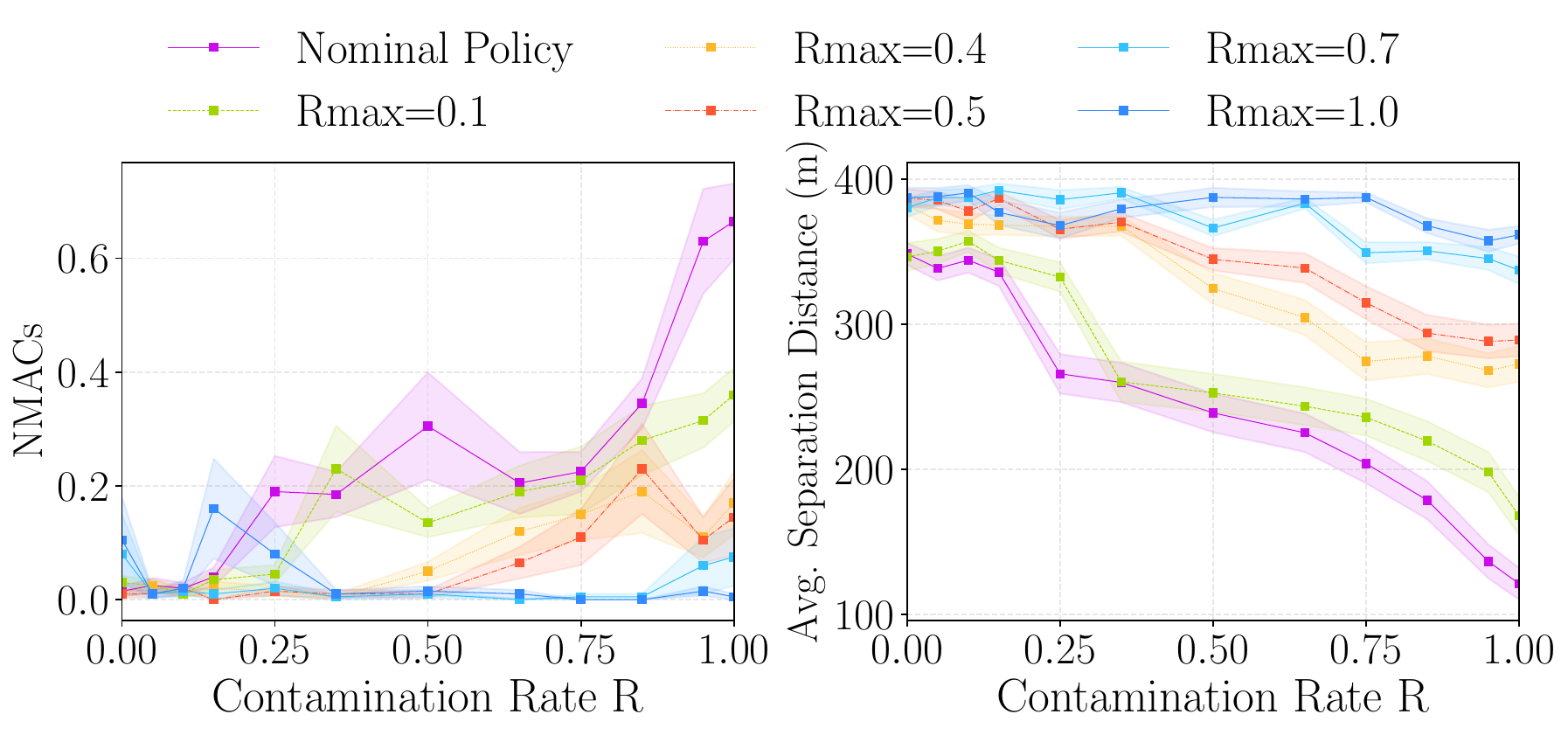}
    \caption{Safety performance of the observation filtering under increasing observation degradation. \emph{Left}: Near mid-air collision (NMAC) count of small UAS. \emph{Right}: Minimum separation distance between the aircraft agents achieved per episode.
    }
    \label{fig:obs_filter}
\end{figure}

Fig. \ref{fig:obs_filter} shows the safety performance of the trained policy when connected with observation filtering, across selected values of assumed uncertainty bounds $R_{\max}$. Unlike action filtering, filtering the observation substantially reduces NMAC rates, but its effectiveness depends on the choice of $R_{\max}$.

Three notable regimes emerge. At $R_{\max}=0.1$ (under-conservative), the filter assumes less uncertainty than is actually present; NMAC rates remain high for $R\ge0.25$, approaching the baseline or nominal policy safety performance. At $R_{\max} = 1.0$ (over-conservative), excessive pessimism introduces false alarms at low GNSS degradation; the NMAC rate at $R=0$ exceeds the nominal policy. At $R_{\max}\approx 0.7$ (balanced), the safety performance is robust across the GNSS degradation levels.
Table \ref{tab:baseline} reinforces this analysis by comparing cases with filtered versus unfiltered observations with $R_{\max}=0.7$ and $\alpha=5$ s. The average NMAC rate across the degradation levels decreased from 0.24 to 0.02 per episode, indicating a reduction of 90\%. Under active GNSS degradation ($R>0$), the NMAC rate experiences a more than 89\% drop. However, when there is no degradation ($R=0$), observation filtering increases the NMAC rate from $0.02$ to $0.08$. 
Furthermore, the separation margins improved as well. The average minimum separation increased from $250$ m to $375$ m, providing a substantial margin above the standard $100$ m. 

\begin{table}[t]
\centering
\caption{NMAC Rate: Unfiltered vs. Filtered Observation with $R_{\max}=0.7$.
\footnotesize $\dagger$\, At $R=0$, observation filtering increases the NMAC rate from 0.02 to 0.08, reflecting the conservatism of worst-case correction under nominal GNSS conditions.
}
\label{tab:baseline}
\begin{tabular}{lccc}
\toprule
$R$ & Unfiltered Obs. & Filtered Obs. & Reduction \\
\midrule
0.00 & 0.02 & 0.08 & $\dagger$ \\
0.25 & 0.19 & 0.02 & 89\% \\
0.35 & 0.19 & 0.01 & 97\% \\
0.50 & 0.31 & 0.01 & 97\% \\
0.75 & 0.23 & 0.01 & 98\% \\
1.00 & 0.67 & 0.08 & 89\% \\
\midrule
Mean & 0.24 & 0.02 & 90\% \\
\bottomrule
\end{tabular}
\end{table}

\subsection{Action Filtering vs Observation Filtering}

\begin{figure}[t]
    \centering
    \includegraphics[width=\linewidth]{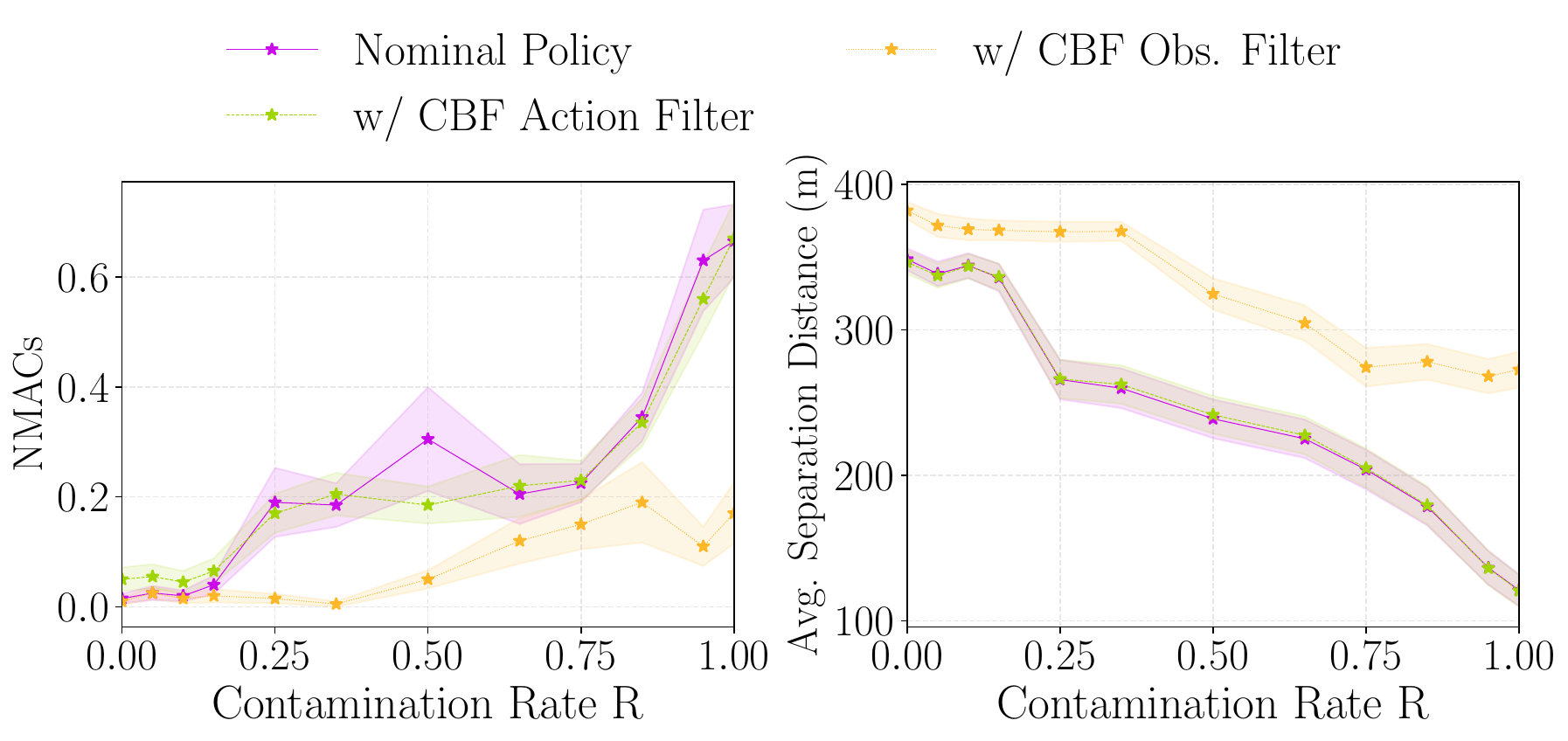}
    \caption{
    Direct comparison of action filtering and observation filtering at $R_{\max} = 0.4$. Observation filtering substantially outperforms both action filtering and nominal operations (without any safeguard).
    }
    \label{fig:why_obs_filtering}
\end{figure}

Comparing the two strategies, as shown in Fig. \ref{fig:why_obs_filtering}, reveals that action filtering (green) tracks the nominal policy (purple), providing little improvement. The observation filter (yellow), on the other hand, dramatically reduces NMAC rates and improves separation distances over GNSS degradation levels.

\subsection{Sensitivity to Velocity Weighting}

The velocity weighting parameter $\alpha$ in \eqref{eq:barrier} intuitively encodes the lookahead horizon for conflict anticipation, for example, between two sUASs. Smaller $\alpha$ (e.g., $1$ s) weights current separation heavily, triggering the intervention only when the aircraft are too close to each other. Larger $\alpha$ (e.g. $10$ s) emphasizes closing-rate, triggering an earlier intervention for fast-approaching traffic even at larger separations. Crucially, the parameter $\alpha$ affects the two filtering approaches differently.

\begin{figure}[t]
    \centering
    \includegraphics[width=\linewidth]{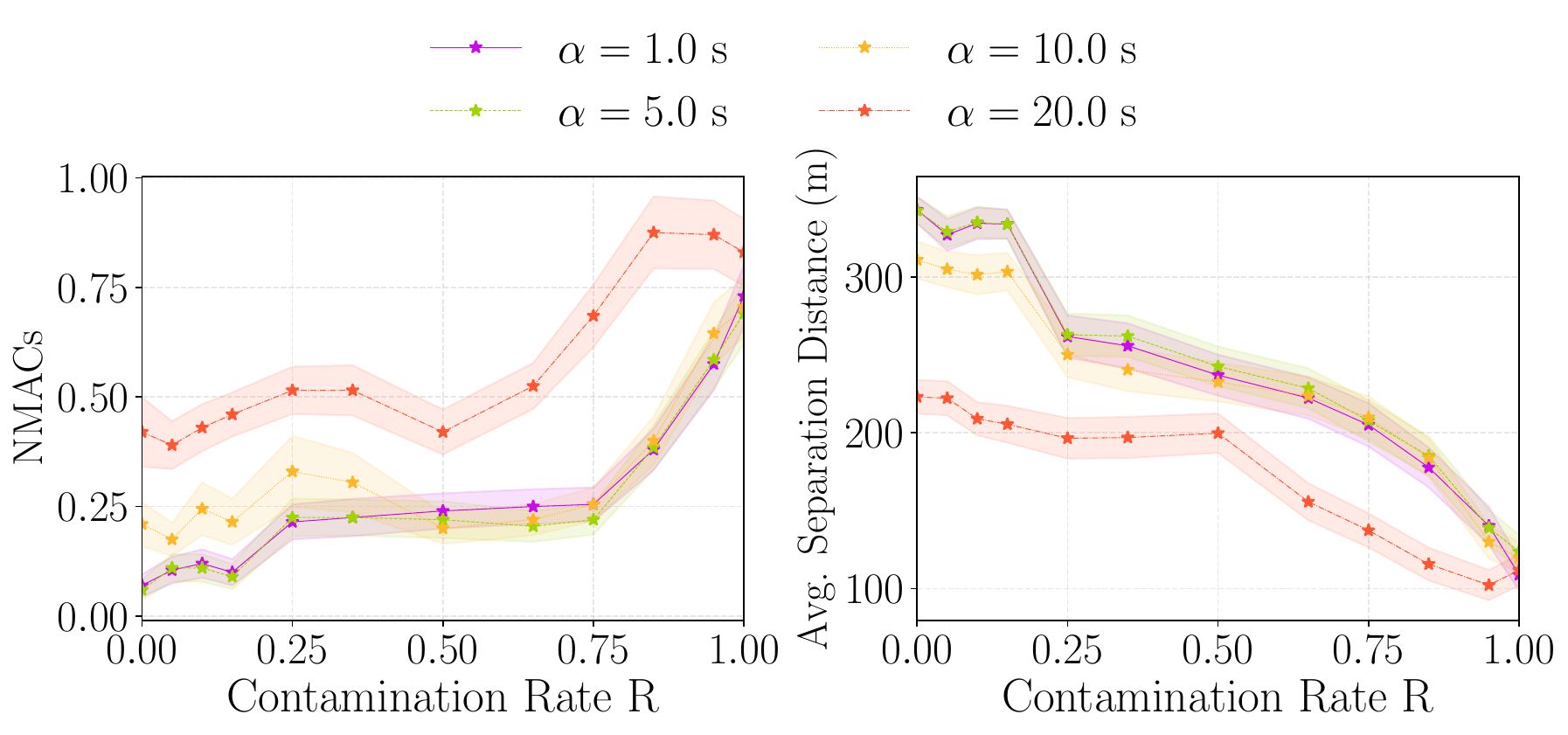}
    \caption{Action filtering is sensitive to velocity weighting $\alpha$ at $R_{\max}=0.7$. Larger $\alpha$ degrades safety: at $\alpha=20$ s, NMAC rates approach baseline levels as aggressive interventions disrupt the tactical conflict resolution sequences that the trained policy has already learned.
    }
    \label{fig:sensitivity_alpha_action_filter}
\end{figure}

Fig. \ref{fig:sensitivity_alpha_action_filter} shows that for action filtering, increasing $\alpha$ degrades safety performance. At $\alpha=1$ s, the NMAC rates remain moderate; at $\alpha=10$ s, they approach baseline levels (nominal policy performance). At $\alpha=20$ s the action filter performs worse than the nominal policy. 

\begin{figure}[t]
    \centering
    \includegraphics[width=\linewidth]{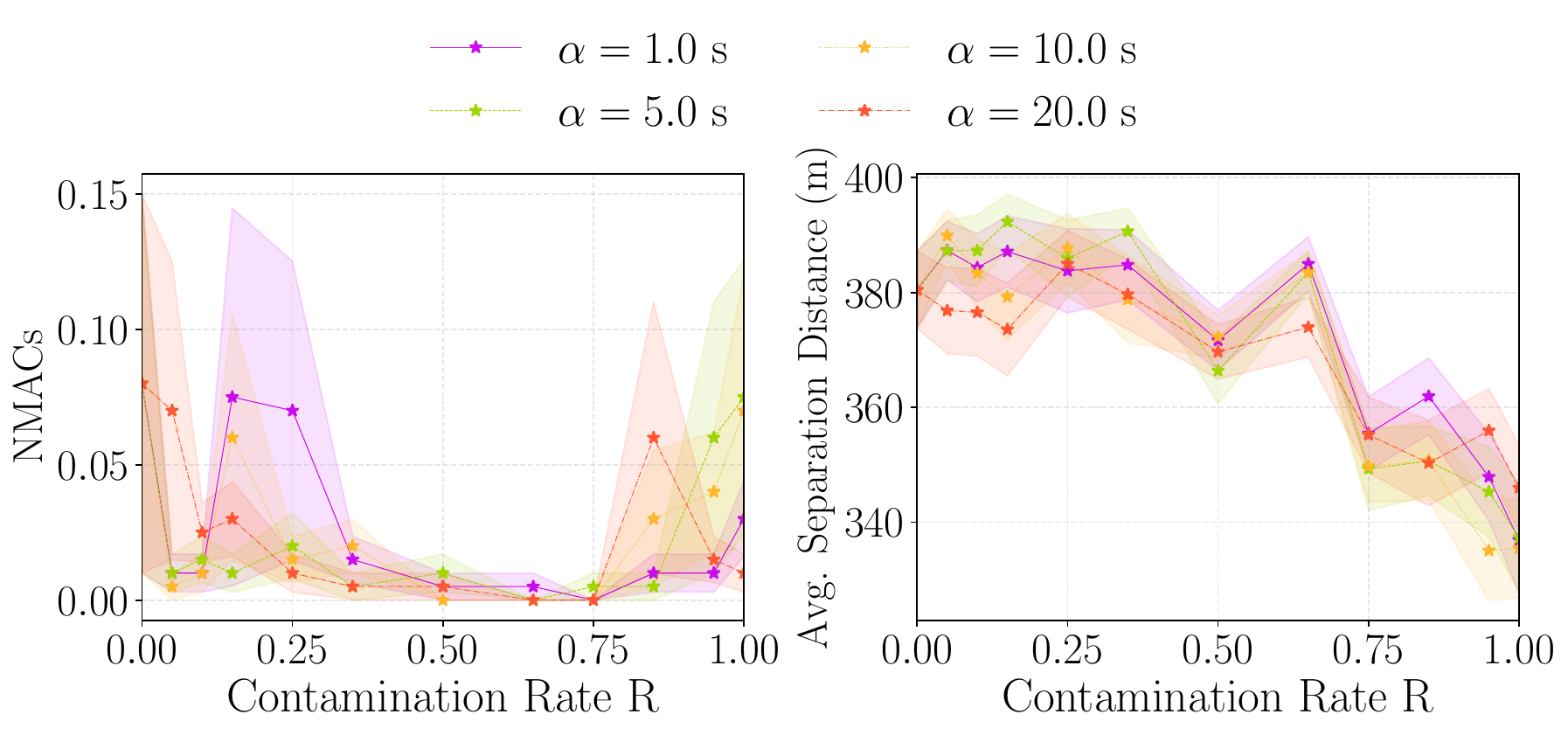}
    \caption{Observation filtering is insensitive to velocity weighting $\alpha$. All settings achieve comparable safety performance at $R_{\max} = 0.7$, indicating that the lookahead horizon encoded in $\alpha$ has secondary importance when positions are conservatively corrected.
    }
    \label{fig:sensitivity_alpha_obs_filter}
\end{figure}

Fig. \ref{fig:sensitivity_alpha_obs_filter} shows the opposite pattern for observation filtering: all settings ($\alpha \in \{1, 5, 10, 20\}$ s) achieve comparable NMAC rates at $R_{\max}=0.7$. The safety performance is effectively invariant to the choice of $\alpha$.

\section{Discussion}
\label{sec:discussion}

\subsection{Why Observation Filtering Outperforms Action Filtering}
\label{sec:discusion_obs_vs_action_filtering}

The asymmetry in our results stems from a mismatch between the safety model that the barrier function encodes and the safety model that the policy has learned. The barrier function evaluates a scalar safety margin: instantaneous separation distance and closing rate, over a single time step. The policy, by contrast, has learned multi-step or lookahead conflict resolution strategies developed through training and reinforcement learning's credit assignment (Section \ref{sec:pretrained_policy}). Action filtering overrides the policy's output using the single-step barrier condition, and in doing so, disrupts these learned resolution strategies. Observation filtering preserves the policy's decision authority, allowing these strategies to operate on corrected inputs.

Intuitively, the discrete-time CBF constraint (\ref{eq:dtcbf}) asks whether the barrier will remain positive in the next step, while the policy's learned behaviors address a different question: will this encounter evolve safely over the horizon relevant to maneuvering dynamics? These questions align in simple geometries but diverge in complex multi-agent scenarios where tactical conflict resolution sequential decisions span multiple time steps. When these questions diverge, action filtering imposes the single-step barrier model while observation filtering defers to the policy's implicit safety \emph{reasoning}, i.e., estimating worst-case observations and trusting the policy's learned responses to these estimations. 

The sensitivity analysis reinforces this distinction. For action filtering (Fig.~\ref{fig:sensitivity_alpha_action_filter}), increasing $\alpha$ amplifies the closing-rate term in the barrier function, causing the CBF constraint to trigger on geometrically benign encounters that the policy would resolve naturally. The resulting interventions, calibrated to a single-step horizon, disrupt the policy's learned multi-step resolution strategies. At $\alpha = 20$ s, small velocity estimation errors propagate into large barrier value changes, making the constraint either too conservative or too permissive, and safety performance falls below that of the unfiltered, nominal policy.
By contrast, observation filtering is insensitive to $\alpha$ (Fig. \ref{fig:sensitivity_alpha_obs_filter}) because the position correction dominates the worst-case estimate. Once each agent is displaced by $\delta_p = R_{\max}\kappa_p$ toward the other, reducing apparent separation by up to $2\delta_p=84$ m at $R_{\max} = 0.7$, the closing-rate term provides only secondary refinement. The policy then responds to the corrected observation using its full learned repertoire, which already encodes appropriate lookahead through training.
The parameter $\alpha$ therefore plays a marginal role: it shapes the worst-case estimate slightly, but the policy's own temporal reasoning subsumes the function that $\alpha$ serves in the barrier constraint.

These results do not imply that CBF-based action filtering is inherently ineffective for multi-agent safety. Recent work on learned barrier functions \cite{cheng2019end} and decentralized neural barrier certificates \cite{qin_2021_learned_cbf} demonstrates safety performance by training the barrier function and policy jointly, allowing both to co-adapt and by learning barrier representations that capture safety-relevant features beyond instantaneous separation and closing rate. Our setting differs in three aspects that compound the limitation: the barrier function is hand-designed rather than learned, it is applied post-hoc to a policy trained without barrier constraints, and the discrete action space limits the filter to coarse corrections. Observation filtering sidesteps all three issues by leaving the policy's decision process intact and intervening only at the input. 

Action filtering also retains one property that observation filtering lacks: it provides an explicit, certifiable forward-invariance guarantee on the filtered action, independent of the policy's behavior. In deployment contexts that demand a formal safety case, for example, certification regimes that require provable constraint satisfaction rather than empirical collision reduction, this guarantee may justify action filtering despite its weaker empirical performance here. Observation filtering, by contrast, inherits its safety properties from the policy and offers no such standalone certificate. 

Overall, this finding extends beyond small UAS separation: for any learned controller with internalized safety behaviors, observation-space interventions may preserve capabilities that action-space constraints would discard.

\subsection{Relation to Training-Time Robustness}
 
Our experiments evaluate observation filtering on a nominally trained policy that has never encountered adversarial perturbations during training. This design isolates the contribution of the runtime mechanism: any safety improvement is attributable to observation filtering rather than training-time robustness.
 
Therefore, a natural extension would evaluate observation filtering on the adversarially robust policy of \cite{zongo_robust_marl}, testing whether runtime filtering provides additional benefit on top of training-time robustness. We conjecture that the two mechanisms are complementary: adversarial training shapes the policy's responses to perceived danger, while observation filtering ensures the policy perceives danger even under degradation. However, observation filtering is not without cost. By presenting the policy with a worst-case view of the traffic state, the filter induces conservative actions, i.e., earlier and more frequent decelerations, even when the true encounter geometry may not warrant that. This conservatism, while beneficial for safety, can increase flight time and reduce operational throughput, particularly at higher values of $R_{\max}$ where the position correction is large. Quantifying this efficiency-safety tradeoff across operational scenarios remains an important direction for future work. 

A separate question is whether the architectural advantage of observation filtering persists when the barrier function itself is learned rather than hand-designed. Existing neural barrier certificate methods \cite{cheng2019end, qin_2021_learned_cbf} assume accurate state information; extending them to certify safety over bounded uncertainty sets, such as those induced by GNSS degradation, would address both the expressiveness limitation identified in Section \ref{sec:discusion_obs_vs_action_filtering} and the observation robustness gap in the current learned-CBF literature.
 
\subsection{Limitations}

Our GNSS degradation model (Section \ref{sec:gnss_degradation}) assumes heading estimates are derived from magnetometer and inertial sensors and remain accurate. This assumption scopes the threat model to position and velocity degradation. However, an adversary capable of perturbing heading observations could defeat the filter by making head-on encounters appear as crossings. Extending the threat model to heading uncertainty would require additional sensing modalities, such as visual odometry, or a tighter integration with inertial navigation.

Moreover, the parameter $R_{\max}$  must bound actual degradation. In fact, underestimating it yields insufficient conservatism, while overestimating it introduces artifacts. Hence, calibrating $R_{\max}$ to the operational GNSS environment, i.e., accounting for urban canyon geometry, expected multipath severity, and spoofing threat level, is essential for deployment. This could mean, for instance, $R_{\max} \approx 0.7-1.0$ for urban canyons or $0.4-0.5$ for suburban. In our case $\alpha$ can be fixed (e.g. at $5$ s), given the observation filtering insensitivity to this parameter. 
 
Furthermore, the approach assumes that the policy generalizes appropriately to worst-case observations. If such states are far outside the training distribution, responses may be unpredictable. Our experiments suggest this is not problematic in practice, likely because worst-case states represent plausible encounter geometries even if they were not explicitly encountered during training.

Finally, our evaluation compares the two filtering architectures against the unfiltered nominal policy rather than against alternative runtime-safety paradigms. Benchmarking observation filtering against Hamilton-Jacobi reachability \cite{fisac2019bridging} filters or learned recovery policies \cite{thananjeyan2021filtering} would further situate its performance within the broader runtime-safety landscape, and we leave this comparison to future work.

\section{Conclusion}
\label{sec:conclusion}

This work compared two runtime safety architectures for learned sUAS separation policies operating under GNSS degradation: action filtering via discrete-time control barrier functions and observation filtering via worst-case state correction. Action filtering provided no measurable safety improvement, while observation filtering reduced near mid-air collisions by 90\% and proved robust to design parameter choices of the hand-crafted barrier function. The decisive factor was not the quality of the worst-case estimate, given that both approaches share identical estimation, but how that estimate was used. Constraining the policy's actions with a single-step barrier condition discarded the multi-step conflict resolution behaviors the policy had internalized through training. Correcting the policy's observations preserved these behaviors, allowing the policy to respond to sensed danger with its full learned capabilities. This finding suggests a broader design principle for composing runtime safety mechanisms with learned controllers: when a policy has internalized safety-relevant behaviors, observation-space interventions that inform the policy may outperform action-space constraints that override the policy outputs.

\end{document}